# Enabling Privacy-Aware AI-Based Ergonomic Analysis

Sander De Coninck[a]*, Emilio Gamba[b], Bart Van Doninck[b], Abdellatif Bey-Temsamani[b],

Sam Leroux[a], Pieter Simoens[a]

[a]IDLab, Department of Information Technology at Ghent University – imec, Technologiepark 126, Ghent B-9052, Belgium
[b]Flanders Make, corelab ProductionS, Gaston Geenslaan 8, 3001 Heverlee, Belgium

* Corresponding author. E-mail address: sander.deconinck@ugent.be

**Abstract**

Musculoskeletal disorders (MSDs) are a leading cause of injury and productivity loss in the manufacturing industry, incurring substantial economic costs. Ergonomic assessments can mitigate these risks by identifying workplace adjustments that improve posture and reduce strain. Camera-based systems offer a non-intrusive, cost-effective method for continuous ergonomic tracking, but they also raise significant privacy concerns. To address this, we propose a privacy-aware ergonomic assessment framework utilizing machine learning techniques. Our approach employs adversarial training to develop a lightweight neural network that obfuscates video data, preserving only the essential information needed for human pose estimation. This obfuscation ensures compatibility with standard pose estimation algorithms, maintaining high accuracy while protecting privacy. The obfuscated video data is transmitted to a central server, where state-of-the-art keypoint detection algorithms extract body landmarks. Using multi-view integration, 3D keypoints are reconstructed and evaluated with the Rapid Entire Body Assessment (REBA) method. Our system provides a secure, effective solution for ergonomic monitoring in industrial environments, addressing both privacy and workplace safety concerns.

*Keywords:* Ergonomic Analysis; Privacy; Human Pose Estimation; Privacy-Aware Machine Learning

## 1. Introduction

In the context of Industry 5.0, companies are prioritizing their employees' well-being, particularly in manufacturing, where poor ergonomics contribute to costly musculoskeletal disorders (MSDs) [1, 2]. With a shortage of skilled labor [3] and the financial impact of MSDs, ergonomic studies on improving workers' physical well-being (and efficiency) by identifying sources of injury risks are gaining popularity.

Continual monitoring by ergonomics experts is time-consuming and expensive, so many look towards automated ways of assessing ergonomic risk. Wearable sensors (e.g., inertial measurement units, force sensors, heart rate monitors) and motion capture systems enable such real-time monitoring, providing insight into the physical strain caused by certain tasks [4]. However, these sensors require the worker to wear equipment, such as gloves or suits, that may be uncomfortable, restrict movement or even interfere with task performance [5].

In contrast, AI-driven camera systems, utilizing machine learning for human pose estimation, offer a non-intrusive alternative for monitoring ergonomic conditions. While these systems provide the advantage of unobtrusive monitoring, they also raise significant privacy concerns, especially regarding their potential misuse for performance tracking. Compliance with the General Data Protection Regulation (GDPR [6]) in the European Union adds another layer of complexity, as companies must navigate the delicate balance between enhancing health and safeguarding privacy while adhering to legal requirements.

To address these concerns, we propose a privacy-aware system for ergonomic assessment. This approach immediately processes captured footage to remove unnecessary information (skin tone, hair color, facial features, ...), leaving only what is required for keypoint estimation. Once transformed, the footage is sent to a central server for ergonomic analysis. This transformation can be executed within a trusted execution environment (TEE)—a secure, isolated part of the camera's CPU—ensuring that personal data is safeguarded. An illustration of our proposed system can be seen in Figure 1.

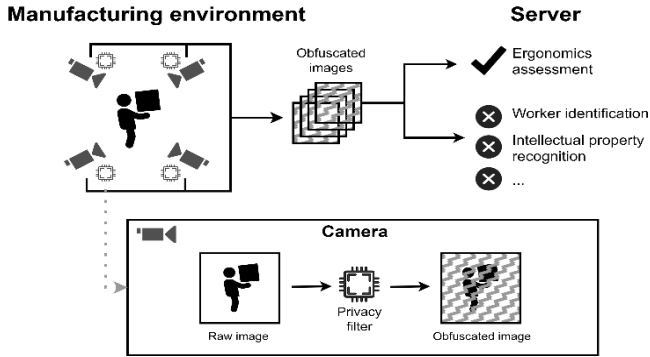

Figure 1: Configuration of our privacy-aware ergonomics system. Camera footage is transformed and obfuscated on the camera, ensuring it is suitable only for keypoint estimation. Once obfuscated, the images are transmitted to a central server for ergonomic assessment

One potential implementation for this privacy-preserving transformation is to blur all individuals detected in the footage [7]. However, this method risks compromising privacy due to potential misdetections, as undetected people will not be blurred. Additionally, sufficient blurring to obscure personal identifiers may hinder its effectiveness for pose estimation. Alternatively, some techniques use inpainting techniques to remove objects completely [8, 9, 10], or replace them with an anonymized alternative [11, 12].

For keypoint estimation, an essential component of ergonomic assessment, some researchers have focused on optimizing the point spread function of the camera lens to capture only necessary data while simultaneously enhancing a neural network for pose estimation [13]. Others have explored alternative sensors, such as wearables and LiDAR data [14]. However, these sensor options are often more expensive or intrusive for employees. Furthermore, specialized hardware can limit compatibility with standard software systems, increasing production costs.

In this paper, we address the privacy challenges associated with ergonomic assessment using computer vision and propose to implement a privacy filter using a Generative Adversarial Privacy (GAP) system [15]. This involves adversarially training an obfuscation and a de-obfuscation network, where one network transforms the data into a version suitable only for keypoint estimation while the other attempts to reconstruct the original image from the obfuscated version. This approach has several advantages. First, we can leverage existing keypoint estimation methods for our obfuscated data. Second, since the adversary focuses solely on reconstructing the original data, there is no need to define undesirable tasks (opt-out). Instead, we adopt an opt-in principle that permits only keypoint estimation, which is inherently more secure against unexpected data leaks. Finally, our technique does not depend on specialized hardware, making it more accessible and cost-effective.

We evaluate the privacy filtering approach in a multi-camera setup, where manufacturing employees are recorded from four camera angles. The privacy filter is applied to the footage from all cameras before processing it through an unmodified pose estimation model. The resulting 2D keypoints are then matched between the different cameras and triangulated to generate 3D keypoints for each pose. The resulting 3D pose is then used to calculate an ergonomic assessment using the REBA score [16], which is the most commonly used assessment method among experts [17]. Our results demonstrate that the technique significantly alters the footage, providing robust privacy protection while minimizing the impact on keypoint estimation accuracy.

The remainder of this paper is structured as follows: In Section 2, we introduce the architecture of our technique. In Section 3 we detail the data collection and ergonomic assessment methods used in our privacy-aware ergonomics use case. The experiments and results are presented in Section 4. Finally, Section 5 summarizes our work and looks at future research directions.

## 2. Privacy-aware ergonomics

To enable privacy-aware ergonomic monitoring in the workplace, we propose the following system: A multi-camera setup is used to capture manufacturing workers from four different angles. Each camera applies a privacy filter to the footage, modifying the original images to produce an opt-in version that restricts the use of the data solely to ergonomic assessments. These transformed frames are then transmitted to a central server, where keypoint detection is performed on each frame. The detected keypoints are then matched and triangulated across the multiple camera views to generate three-dimensional coordinates. These 3D keypoints are then utilized to calculate a REBA score according to the method proposed in [16], which provides an ergonomic assessment of the situation. Figure 2 illustrates our system workflow.

### 2.1. Generative adversarial training

To facilitate this privacy transformation, we implement an autoencoder neural network proposed by De Coninck et al. [18]. This network is trained to optimize the accuracy of a pose estimation model on its output while simultaneously preventing a separate autoencoder, known as the deobfuscator, from reconstructing the original images from its output. In this manner, the obfuscator is engineered to retain only the essential information necessary for pose recognition while limiting access to additional data. For our pose detection component, we utilize the state-of-the-art model YOLOv8-Pose [19].

The obfuscator $O$ is trained with the following loss, with $X$ being the input image, $\alpha$ a weighting factor and $L_{pose}$ the original YOLO pose estimation loss [19]:

$$L_{obf}(X) = L_{pose}\big(O(X)\big) - \alpha L_{deobf}\big(X, O(X)\big) \qquad (1)$$

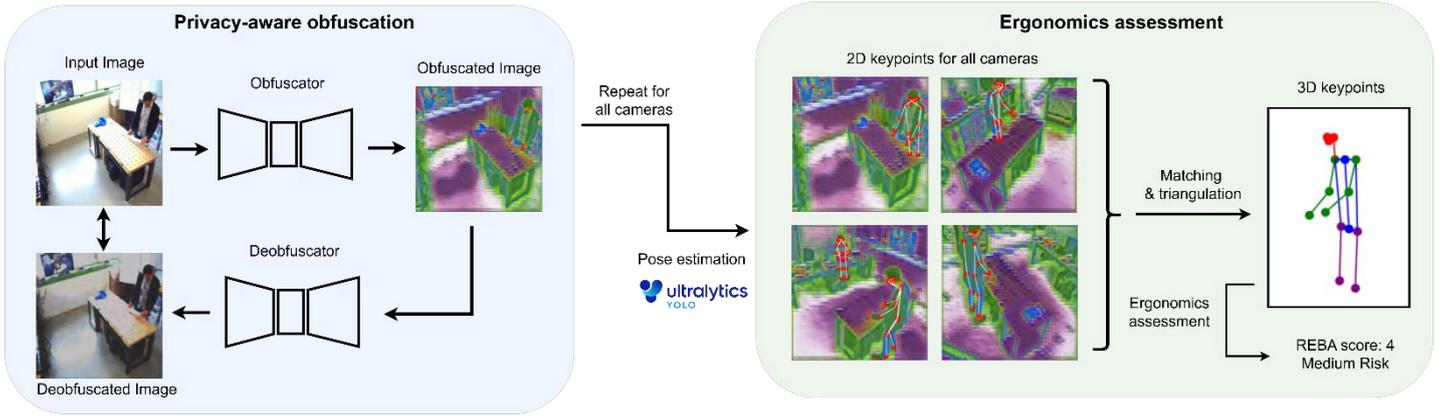

Figure 2: Workflow of our privacy-aware ergonomics assessment system. The images are obfuscated using a Generative Adversarial Privacy scheme as described in [18]. This obfuscation process is applied to each camera, and the resulting anonymized frames are used to detect 2D keypoints with an unmodified YOLO model. After matching keypoints across views and performing triangulation, a 3D representation of the worker is constructed, enabling the calculation of a REBA score.

The training objective for the deobfuscator $D$ is to minimize the loss $L_{deobf}$ defined as follows:

$$L_{deobf}(X, O(X)) = (X - D(O(X)))^2 \qquad (2)$$

We pre-train all models on the COCO-keypoints dataset [20] for 20 epochs with batch size 8. Both the obfuscator and deobfuscator are optimized using the AdamW optimizer [21] with an initial learning rate of $1 \times 10^{-3}$, which is decreased by a factor of 10 every 10 epochs. The training was conducted on a Tesla V100 GPU with automatic mixed precision to enhance efficiency. Following this pre-training phase, we finetune the models on our dataset for an additional 6 epochs using a learning rate of $1 \times 10^{-4}$. After training, the deobfuscator is discarded, and only the obfuscator is used for inference.

### 2.2. Evaluation metrics

We evaluate our approach by assessing its effectiveness in enabling ergonomic evaluations while also measuring its ability to protect privacy.

#### 2.2.1. Ergonomic evaluation

We measure the capability to do ergonomic evaluations through keypoint estimation metrics and REBA calculation capacity. In human keypoint estimation tasks, Average Precision (AP) is commonly used to evaluate the performance of a model. The AP is calculated by measuring the overlap between the predicted and ground truth keypoints using the Object Keypoint Similarity (OKS) [20] metric. The OKS serves a role similar to the Intersection over Union (IoU) used in object detection. For each object, ground truth keypoints are given as $[x_1, y_1, v_1, ..., x_k, y_k, v_k]$, where $(x_i, y_i)$ are the keypoint coordinates, and $v_i$ is a visibility flag. The object's scale $s$ is defined as the square root of the object's segment area. Predicted keypoints have the same format but do not require visibility prediction.

The Object Keypoint Similarity (OKS) is calculated as:

$$\text{OKS} = \frac{\sum_i \left[\exp\left(-\frac{d_i^2}{2s^2\kappa_i^2}\right)\delta(v_i > 0)\right]}{\sum_i \delta(v_i > 0)} \qquad (3)$$

where $d_i$ is the Euclidean distance between the predicted and ground truth keypoints, $\kappa_i$ is a keypoint-specific scale factor, and $\delta(v_i > 0)$ is an indicator function that includes only visible or labeled keypoints in the calculation.

The REBA score is one of the most used ergonomic assessment scores among experts [17]. It evaluates the risk of a given pose based on the body's joint angles, such as the elbow angle formed by the forearm and the upper arm, taking into account task-specific factors such as repetitive movements, load handling, and force exertion. It calculates risk scores for various body parts, including the neck, trunk, legs, upper and lower arms, and wrists. These individual scores are then combined to generate an overall risk score. We refer to [16] for a more detailed explanation.

#### 2.2.2. Privacy protection

For privacy assessment, we utilize image similarity metrics to evaluate the level of degradation from the original image. Common metrics include SSIM [22], PSNR, VIF [23], and LPIPS [24]. However, these metrics do not always correspond to human perception [25]. To address this, we prioritize the SemSim [25] metric for privacy evaluation, as it more closely aligns with human judgment. This metric is based on a model trained using human-annotated ground truth values for image pairs. The SemSim score is then computed by measuring the Euclidean distance in the feature space of the trained model.

### 3. Case study: privacy-aware ergonomics in a manufacturing environment

In this case study, we record human operators performing a simple assembly task. The objectives of this case study are twofold. First, we aim to evaluate how well privacy can be preserved by obfuscating human operators performing the assembly. Second, we aim to assess the accuracy of the 3D keypoints reconstructed from the obfuscated video data to provide an accurate ergonomic evaluation using the REBA method.

*3.1. Data collection*

We instructed operators to build a frame by connecting aluminum profiles to angle brackets in a work cell equipped with a table (at a height of 86cm), a tool cabinet, and components stored in a cabinet with drawers at knee height.

The operators were asked to perform the assembly on the table, meaning that some steps required them to adopt high-risk postures according to the REBA scoring, such as fetching components from drawers, bending over a table, and picking up tools placed on the floor. An example of such a situation can be seen in Figure 3.

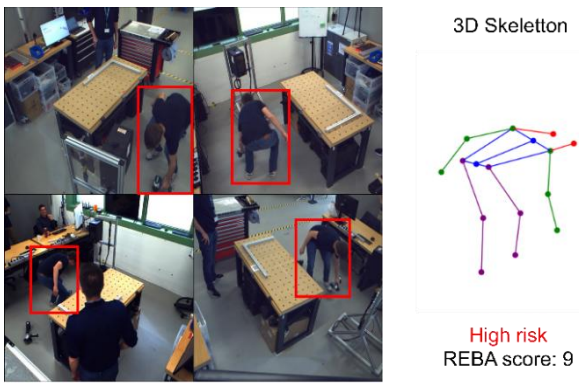

Figure 3: Example risk situation with constructed 3D skeleton and REBA analysis

The operators were filmed by 4 x 5 MP RGB[1] cameras capturing each person from different angles as shown in Figure 3. During the nine assembly tasks performed by individuals or teams of two, a total of 20 different people were recorded, including bystanders and passers-by. We used data from seven tasks to train our obfuscation model, reserving the remaining two for testing.

*3.2. Analysis*

We performed keypoint detection on all frames from the test set and performed the REBA analysis on the resulting 3D keypoints (as seen in Figure 3). The overall distribution of scores, along with their corresponding risk labels, calculated on two assembly tasks is shown in Figure 4. It reveals that all risk categories are represented, with low and medium risk being the most prevalent. High-risk situations are also captured, though they occur less frequently than the other categories

*Limitations.* The participants in this exploratory case study were colleagues who volunteered for the task recording, which may not fully capture the diversity of postures in a typical industrial setting. Additionally, as the primary focus of this paper is on privacy, ergonomics aspects such as repetitive motions, load handling, and forces exerted were not explored. Therefore, REBA scores will never exceed 10.

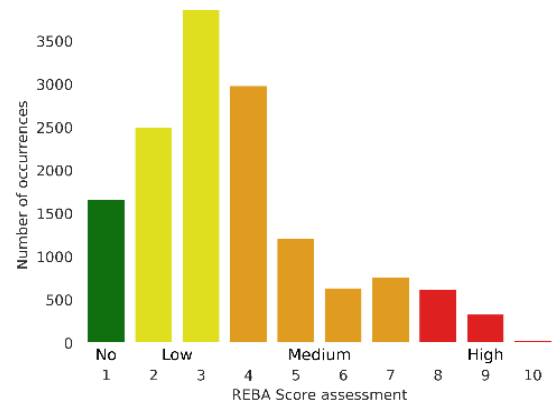

Figure 4: Distribution of REBA scores and risk categories on two assembly tasks

**4. Experiments, results and discussion**

In this section, we present the results of our obfuscation technique and evaluate its effectiveness in protecting privacy, using both qualitative and quantitative approaches. We also compare ourselves to different obfuscation techniques in terms of allowing ergonomics vs privacy protection. Finally, we assess the downstream performance of ergonomic evaluations using REBA scores, which are calculated from 3D keypoints estimated from obfuscated frames. The results are based on a held-out set of two manufacturing tasks, totaling approximately 10 minutes of footage captured from four different angles.

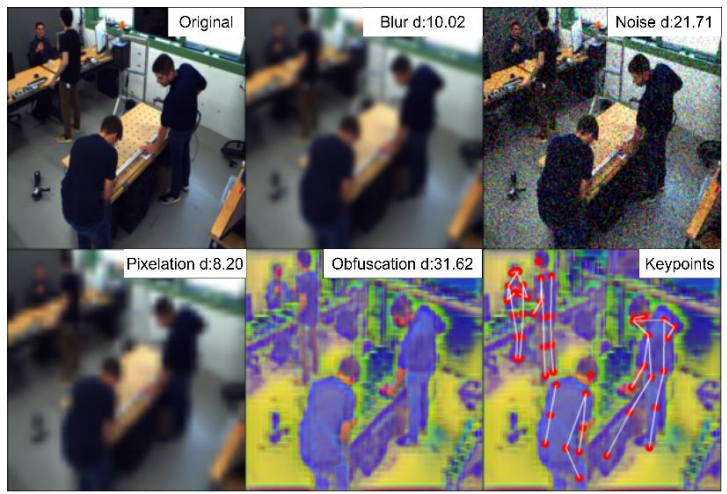

Figure 3: Visualisation of images transformed with blur, additive noise, pixelation and our obfuscation technique, shown together with the SemSim distance (denoted as d). The last frame shows the keypoints detected on our obfuscated frames

---

[1] Each Daheng MER2-503-23-GC-P camera is fitted with a Daheng LCM-5MP-08MM-F1.4-1.5-ND1 lens

*4.1. Visual analysis of obfuscated frames*

We show some visual results or our obfuscation, the calculated SemSim distances, and classic obfuscation techniques in Figure 5. The obfuscated images retain information necessary to do pose estimation but remove any superfluous information. Visually, we see that people are still recognizable as people, which would be necessary to perform keypoint detection. However, the number of identifiable factors is greatly diminished, details required to recognize faces, as well as of the environment, have not been retained.

*4.2. Ergonomics assessment vs privacy protection*

Next, we demonstrate the tradeoff between privacy and utility by plotting the person mean AP (at a threshold of 50% OKS) for keypoint estimation alongside the SemSim distance between the original and obfuscated images. We compare our technique with basic obfuscation methods: gaussian blur, additive noise, and pixelation at varying intensities. The results of this experiment are shown in Figure 6. Our obfuscation method achieves the highest SemSim distance, indicating the strongest level of obfuscation, while maintaining approximately 75% mAP50. In contrast, traditional obfuscation techniques show a much worse tradeoff, where stronger obfuscation leads to a significant decline in keypoint estimation accuracy.

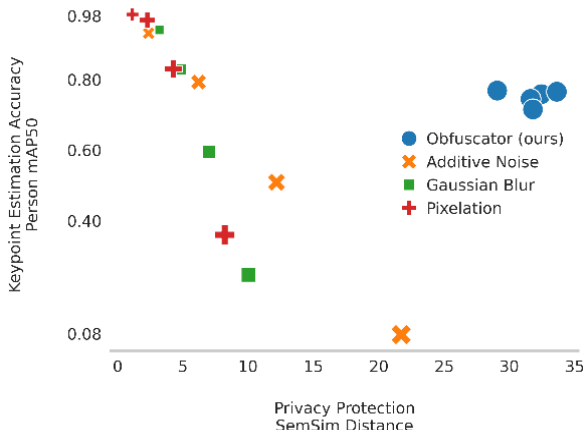

Figure 6: Comparison of the privacy-ergonomics trade-off of our technique versus common obfuscation techniques. We find our technique achieves an optimal balance between privacy and utility.

*4.3. REBA assessment on obfuscated videos*

Finally, we evaluate the ergonomic assessments conducted on obfuscated videos. While keypoint estimation accuracy can serve as an indicator, there are additional considerations since REBA score calculation depends on accurate detection from multiple camera angles. Even small changes in keypoint location can impact the resulting 3D skeleton. However, the involvement of multiple views could also introduce some flexibility if a keypoint is not well detected in one angle. Figure 7 shows the results of this comparison, including the distribution of REBA scores and categories for both the original and obfuscated videos. Note that in 14% of cases, no match was found in the obfuscated frames; this figure only reports where there was. In most cases, the REBA scores from the original and obfuscated videos are very similar, either matching exactly or falling within the same risk category. The assessments tend to be most accurate in lower-risk scenarios, while higher-risk scenarios being both less common and showing larger differences between original and obfuscated videos pose more challenges. This discrepancy can be attributed to the lower frequency of high-risk situations in the training data, resulting in fewer instances for the obfuscator to learn from.

In Figure 8, we present the differences in partial scores used in the REBA calculation. We find that, for most cases, the difference is 0. However, it is noteworthy that the neck shows a higher likelihood of score differences, possibly due to the fine-grained nature of the keypoints in this region. Additionally, the obfuscated score tends to be higher for some body regions, such as the legs, while for the trunk, the opposite trend is more common.

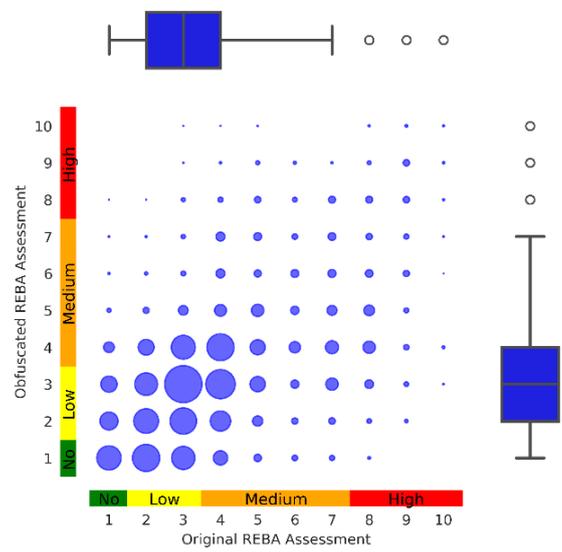

Figure 7: Comparison of REBA scores and risk categories calculated on original and obfuscated data. Generally, REBA scores are similar between obfuscated and original and fall in the same category. For high-risk situations, the discrepancy becomes larger

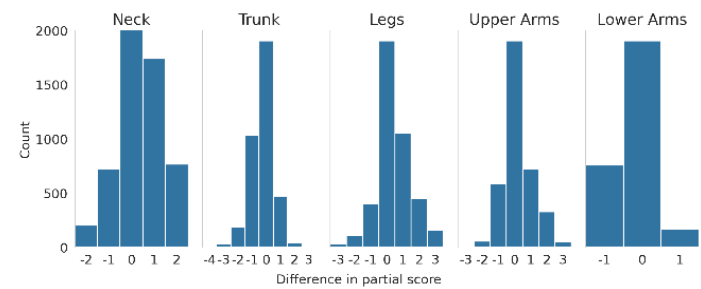

Figure 8: Difference between original and obfuscated partial scores

## 5. Conclusion & future work

In this paper, we presented a privacy-aware system for ergonomic assessment in manufacturing environments, addressing the challenge of balancing employee well-being with privacy concerns in the context of Industry 5.0. By leveraging a Generative Adversarial Privacy (GAP) system, we obfuscated video footage while maintaining the accuracy required for keypoint estimation and ergonomic evaluation using the REBA method. Our system processes video data in a secure manner, transforming it into a format suitable only for keypoint analysis, thus safeguarding sensitive personal information.

Through our experiments on a multi-camera setup, we demonstrated that our privacy filtering technique significantly obfuscates visual data, ensuring robust privacy protection. At the same time, it retains sufficient accuracy for keypoint detection and ergonomic assessment, achieving a more optimal privacy-utility tradeoff compared to traditional obfuscation methods like Gaussian blur and pixelation. The results showed that the REBA scores calculated from the obfuscated videos closely align with those from the original footage, particularly in low-risk scenarios, which are more common in typical manufacturing tasks.

Despite the promising results, our system has some limitations. It tends to underperform in high-risk situations, where precise keypoint estimation is crucial for accurate ergonomic assessment. We have not yet evaluated its suitability for edge devices, which is important for real-time, privacy-preserving processing in industrial settings. We also tested only one keypoint estimation model, leaving room to explore alternative models that may improve accuracy and robustness. Moreover, our privacy evaluation is limited to image degradation metrics. Future work should assess whether obfuscated frames can still be used for privacy-invading tasks, such as person identification or recognition of sensitive attributes. Addressing these limitations will be key to improving the system's real-world applicability.

**Acknowledgements**

Sander De Coninck receives funding from the Special Research Fund of Ghent University under grant no. BOF22/DOC/093. This research is done in the framework of the Flanders AI Research Program (https://www.flandersairesearch.be/en). That is financed by EWI (Economie Wetenschap & Innovatie), and Flanders Make (https://www.flandersmake.be/en), the strategic research Centre for the Manufacturing Industry who owns the Operator 4.0/5.0 infrastructure